\documentclass[preprint,12pt]{elsarticle}
\usepackage{graphicx,subcaption}
\usepackage{lineno,hyperref}
\usepackage{natbib}
\usepackage{geometry}
\usepackage{fleqn}
\usepackage{graphicx}
\usepackage{newtxtext,newtxmath}
\usepackage{hyperref}
\usepackage{booktabs}
\usepackage{bm}
\usepackage{amsmath}
\usepackage{enumitem}
\usepackage{esvect}
\usepackage{soul,xcolor}
\modulolinenumbers[5]

\graphicspath{{./figures/}}
\journal{Elsevier}

\begin{document}

\begin{frontmatter}

\title{\large{An Investigation on Machine Learning Predictive Accuracy Improvement and Uncertainty Reduction using VAE-based Data Augmentation
}}

\author[NCSU]{Farah Alsafadi}
\author[NCSU]{Mahmoud Yaseen}
\author[NCSU]{Xu Wu\corref{mycorrespondingauthor}}
\cortext[mycorrespondingauthor]{Corresponding author}
\ead{xwu27@ncsu.edu}

\address[NCSU]{Department of Nuclear Engineering, North Carolina State University    \\ 
	Burlington Engineering Laboratories, 2500 Stinson Drive, Raleigh, NC 27695 \\}

\begin{abstract}

    The confluence of ultrafast computers with large memory, rapid progress in Machine Learning (ML) algorithms, and the availability of large datasets place multiple engineering fields at the threshold of dramatic progress. However, a unique challenge in nuclear engineering is \textit{data scarcity} because experimentation on nuclear systems is usually more expensive and time-consuming than most other disciplines. One potential way to resolve the data scarcity issue is deep generative learning, which uses certain ML models to learn the underlying distribution of existing data and generate synthetic samples that resemble the real data. In this way, one can significantly expand the dataset to train more accurate predictive ML models. In this study, our objective is to evaluate the effectiveness of data augmentation using variational autoencoder (VAE)-based deep generative models. We investigated whether the data augmentation leads to improved accuracy in the predictions of a deep neural network (DNN) model trained using the augmented data. Additionally, the DNN prediction uncertainties are quantified using Bayesian Neural Networks (BNN) and conformal prediction (CP) to assess the impact on predictive uncertainty reduction. To test the proposed methodology, we used TRACE simulations of steady-state void fraction data based on the NUPEC Boiling Water Reactor Full-size Fine-mesh Bundle Test (BFBT) benchmark. We found that augmenting the training dataset using VAEs has improved the DNN model's predictive accuracy, improved the prediction confidence intervals, and reduced the prediction uncertainties.

\end{abstract}


\end{frontmatter}


\newpage
\section{Introduction}

The confluence of ultrafast computers with large memory, rapid progress in Machine Learning (ML) algorithms, and the availability of large datasets place multiple engineering fields at the threshold of dramatic progress. However, a unique challenge in nuclear engineering is \textit{data scarcity} because experimentation on nuclear systems is usually more expensive and time-consuming than most other disciplines. Large amounts of data may be available for certain parts such as pipes, pumps and turbines, etc., due to large network of sensors, but not for many others, such as critical heat flux in thermal-hydraulics experiments, advanced materials qualification data like molten salts and multi-principal element alloys, etc. Particularly concerning is the lack of data for advanced reactor design and safety analysis, raising challenges for utilizing ML in licensing analyses of advanced nuclear reactors. In these cases, we need to move beyond ``throw more data and re-train'' at the problem, which is the common solution in areas such as computer vision and natural language processing that have access to ``big data''.

One potential way to address the data scarcity issue is \textit{data augmentation using deep generative learning}. Deep generative learning is an unsupervised ML technique that aims at discovering and learning the regularities or patterns in existing data using \textit{deep generative models (DGMs)}, in order to generate new samples that plausibly could have been drawn from the real dataset. DGMs are typically neural networks (NNs) trained to learn or approximate the underlying distribution of the training data. This enables them to generate synthetic samples that closely match the distribution of the original training data. By employing DGMs for data augmentation, one can significantly expand the training dataset for ML models to achieve better performance in other tasks, such as data-driven predictive ML models. 

Data augmentation with DGMs is still a relatively new research area in nuclear engineering, but has been studied for a few years in computer vision and natural language processing for datasets involving images, audios, videos, spoken words, etc. One widely adopted DGM is the variational autoencoders (VAEs) model \cite{kingma2013auto}. The VAE model utilizes autoencoders to learn the underlying probabilistic distribution of a training dataset. Its primary objective is to approximate the distribution of the training data in order to generate new samples. It has proven successful in data augmentation across various domains, including acoustic modeling \cite{nishizaki2017data} and clinical studies \cite{papadopoulos2023variational}. Additionally, VAEs demonstrate their effectiveness in enhancing product quality prediction through the generation of artificial quality values for training \cite{lee2023developing} and they have been more frequently applied in computer vision tasks, such as generating static images \cite{walker2016uncertain}.

As for applications in engineering, VAEs were used to generate temperature, velocity, and species mass fraction predictions within a computational fluid dynamics (CFD) data-driven surrogate model, allowing for predicting CFD data fields with reasonable accuracy \cite{laubscher2020application}. In \cite{wang2021flow}, the authors introduced an artificial neural network-VAE (ANN-VAE) model for the reconstruction of three-dimensional flow fields, significantly reducing the computational cost of field predictions. In nuclear engineering, a convolutional variational autoencoding gradient-penalty Wasserstein generative adversarial network with random forest (CVGR) was proposed to mitigate imbalance data problem in fault diagnosis of nuclear power plants \cite{guo2024imbalanced}.

In a previous work \cite{alsafadi2023deepNED}, we have developed and compared DGMs based on generative adversarial networks (GANs), normalizing flows (NFs), VAEs, and conditional VAEs for synthetic data generation. It was discovered that these deep generative models have a great potential to augment scientific data, not only image data as studied in the majority of their previous applications. In another work \cite{alsafadi2024predicting}, we utilized conditional VAEs for data augmentation of a critical heat flux dataset. The study included uncertainty quantification and domain generalization analyses, showing that conditional VAEs have a good potential in the nuclear engineering field.

This work will report some new progress on investigating the impact of data augmentation on DNN prediction accuracy, as well as on the reduction of DNN prediction uncertainties. We will employ the VAEs model because it has demonstrated superior performance than other tested DGMs. Our primary objectives are to quantitatively investigate the benefits of data augmentation on deep neural network (DNN) predictive performance by checking (1) if the prediction accuracy can be improved, (2) whether confidence intervals (CIs) of DNN predictions can be narrowed, and (3) whether the prediction uncertainty can be reduced. 
The first objective was achieved by comparing the DNN's predictions using both the original and augmented datasets with different numbers of synthetic samples, and conducting a thorough comparison of data augmentation effects on model performance by utilizing different evaluation metrics.

The second objective was achieved using the conformal prediction (CP) method to compute CIs for the DNN predictions. CP was introduced in \cite{vovk2005algorithmic} as a novel methodology for computing prediction sets for deterministic models. It has been widely utilized across various scientific fields due to its model-agnostic nature, making it applicable to any regression or classification model. In \cite{najera2024uncertainty}, the authors investigate the use of conformal sets for quantifying uncertainty in a deterministic structure-preserving neural network and its implementation within a structural health monitoring framework. In \cite{portela2024conformal}, two variants of CP were proposed, demonstrating promising results across diverse biological data structures and scenarios, offering a general framework for uncertainty quantification in dynamic models of biological systems. In \cite{angelopoulos2020uncertainty}, a modified version of CP was applied to convolutional image classifiers and tested on the ImageNet and ImageNet-V2 datasets using ResNet-152 models. The study in \cite{eliades2017conformal} explores the application of the CP framework to provide reliable confidence information for automatic face recognition.

To accomplish the third objective in this study, we leveraged a Bayesian neural network (BNN) \cite{ghahramani2016history} to capture the uncertainties in the DNN's predictions. The BNN employs Bayesian inference to quantify the statistical distributions of the DNN parameters (e.g., weights), rather than deterministic values by optimization (e.g., stochastic gradient descent). When making predictions, the BNN model generates many samples for network parameters based on their posterior distributions, which are used to produce an ensemble of outputs given the same input. Subsequently, we compute both the mean value and standard deviation of the output samples, which represent the mean prediction and the uncertainty, respectively. Our investigation focuses on assessing the impact of data augmentation on both prediction accuracy and uncertainty by employing uncertainty statistics and predictive likelihood as our evaluation metrics.

The remainder of the paper is structured as follows: In Section \ref{section:Problem-Definition}, we define the problem by providing a brief description of the demonstration example. In Section \ref{section:Methodologies}, we present the methodologies used in this study, including VAE, CP and BNN. The results of our investigation, including the evaluation metrics for both predictive accuracy and associated uncertainty, are presented in Section \ref{section:Results}. Finally, Section \ref{section:Conclusions} discusses the findings and provides our concluding remarks.

\section{Problem Definition}
\label{section:Problem-Definition}

Similar to the previous work \cite{alsafadi2023deepNED}, we will use TRACE simulations of steady-state void fraction data based on the NUPEC Boiling Water Reactor Full-size Fine-mesh Bundle Test (BFBT) benchmark \cite{neykov2005nupec} as the training dataset. More details about the BFBT benchmark, TRACE simulation setup and data augmentation problem definition can be found in \cite{alsafadi2023deepNED}. In this section, we will provide a high level introduction to the problem definition in order to make this paper self-contained.

The BFBT facility replicates a full-scale boiling water reactor fuel assembly. This facility conducts void fraction distribution measurements at four specific axial positions, utilizing X-ray CT scanners and X-ray densitometers. These measurements yield void fraction values labeled as \texttt{VoidF1}, \texttt{VoidF2}, \texttt{VoidF3} and \texttt{VoidF4}, from the bottom to the top. The void fraction distributions were measured across five distinct assembly configurations, denoted as types 0 to 4, each characterized by unique geometries and power profiles. Within each assembly configuration, multiple tests were performed with different design variables including pressure, mass flow rate, power, and inlet temperature. For this work, we have chosen assembly 4101 test number 55 because this test has been used to provide a baseline TRACE input deck in the BFBT benchmark. 

The training dataset is produced by running TRACE code version 5.0 Patch 4. TRACE offers the capability to introduce perturbations to 36 physical model parameters (PMPs), as detailed in the TRACE manual \cite{USNRC2014TRACE}. In a prior study, sensitivity analysis was employed to identify the most influential physical parameters for the BFBT benchmark \cite{wu2018inverse}. From this analysis, five important parameters were selected as listed in Table \ref{table:TRACE-model-parameters}. These five PMPs serve as the \textit{inputs} of the training dataset, with the four void fraction responses as the \textit{outputs}.

\begin{table}[htbp]
	\footnotesize
	\captionsetup{justification=centering}
	\caption{List of TRACE PMPs (multiplicative factors) that are significant to the BFBT benchmark \cite{wu2018inverse}.}
	\label{table:TRACE-model-parameters}
	\centering
	\begin{tabular}{l c c c c}
		\toprule
		Physical model parameters in TRACE & Symbols & Uniform ranges & Nominal values\\
		\midrule
		Single phase liquid to wall heat transfer coefficient& \texttt{P1008} & (0.0, 5.0) & 1.0\\
		Subcooled boiling heat transfer coefficient& \texttt{P1012}   & (0.0, 5.0) & 1.0\\
		Wall drag coefficient & \texttt{P1022} & (0.0, 5.0) & 1.0\\
		Interfacial drag (bubbly/slug Rod Bundle) coefficient & \texttt{P1028} & (0.0, 5.0) & 1.0\\
		Interfacial drag (bubbly/slug Vessel) coefficient & \texttt{P1029} & (0.0, 5.0) & 1.0\\
		\bottomrule
	\end{tabular}
\end{table}

To generate the training dataset, we ran the TRACE code 200 times while randomly perturbing these five PMPs within their uniform ranges listed in Table \ref{table:TRACE-model-parameters}. Consequently, each of the 200 samples in the training dataset is a nine-dimensional vector with five PMPs as inputs and four void fractions as outputs. This data will be referred to as \textit{original training data}. It was also used in our previous work \cite{alsafadi2023deepNED} to train the VAE model. Similarly, we generated a testing dataset from 200 additional TRACE simulations to evaluate the performance of the DNN models on unseen data. Utilizing the trained VAE model, 500 synthetic samples were generated and validated through comparison with TRACE results. In this work we will use these 500 synthetic samples to expand the size of the original training dataset.

To investigate whether expanding the training dataset size with the synthetic samples will lead to improved DNN predictions, our approach involves first training a DNN model using the original training dataset, referred to as the \textit{baseline DNN model}. Then, We will  train a sequence of additional DNN models, each time increasing the training dataset size by adding 100 synthetic samples. The prediction accuracy of these models will then be compared to that of the baseline DNN model, which is trained without data augmentation. Subsequently, we will use the CP method to compute CIs of the DNN predictions and analyze how their widths change as more synthetic data are added. Additionally, we will assess the uncertainties in the DNN predictions derived from different training datasets using BNN.

\section{Methodologies}
\label{section:Methodologies}

\subsection{VAEs for Synthetic Data Generation}

The VAE model is a powerful class of DGM that utilizes autoencoders to learn the data distribution. Its primary objective is to approximate the distribution of the training data in order to generate new samples. It comprises three key elements: the \textit{encoder}, the \textit{latent space} (also referred to as the \textit{bottleneck}), and the \textit{decoder}. However, VAEs are different from the conventional autoencoders in the encoding process. In autoencoders, the encoder's role is to encode the input data into the latent space while preserving the most significant features. Subsequently, the decoder uses this encoded data to recreate the original input by reversing the compression process, effectively transitioning from the latent space back to the initial space. This can be useful in several tasks such as image reconstruction \cite{han2022inference}, denoising \cite{Liu_2020_CVPR_Workshops} and anomaly detection \cite{GONZALEZMUNIZ2022108065}, yet they are not suitable for data generation. Autoencoders are trained to minimize reconstruction loss without taking into account the structure of the latent space. Consequently, if we intend to employ the decoder as a generator for generating new samples, it becomes challenging to establish suitable values for the latent space in the generation phase. If we were to employ a random vector as input, the resulting output would lack a meaningful interpretation.

\begin{figure}[!ht]
	\centering
	\captionsetup{justification=centering}
	\includegraphics[width=0.7\textwidth]{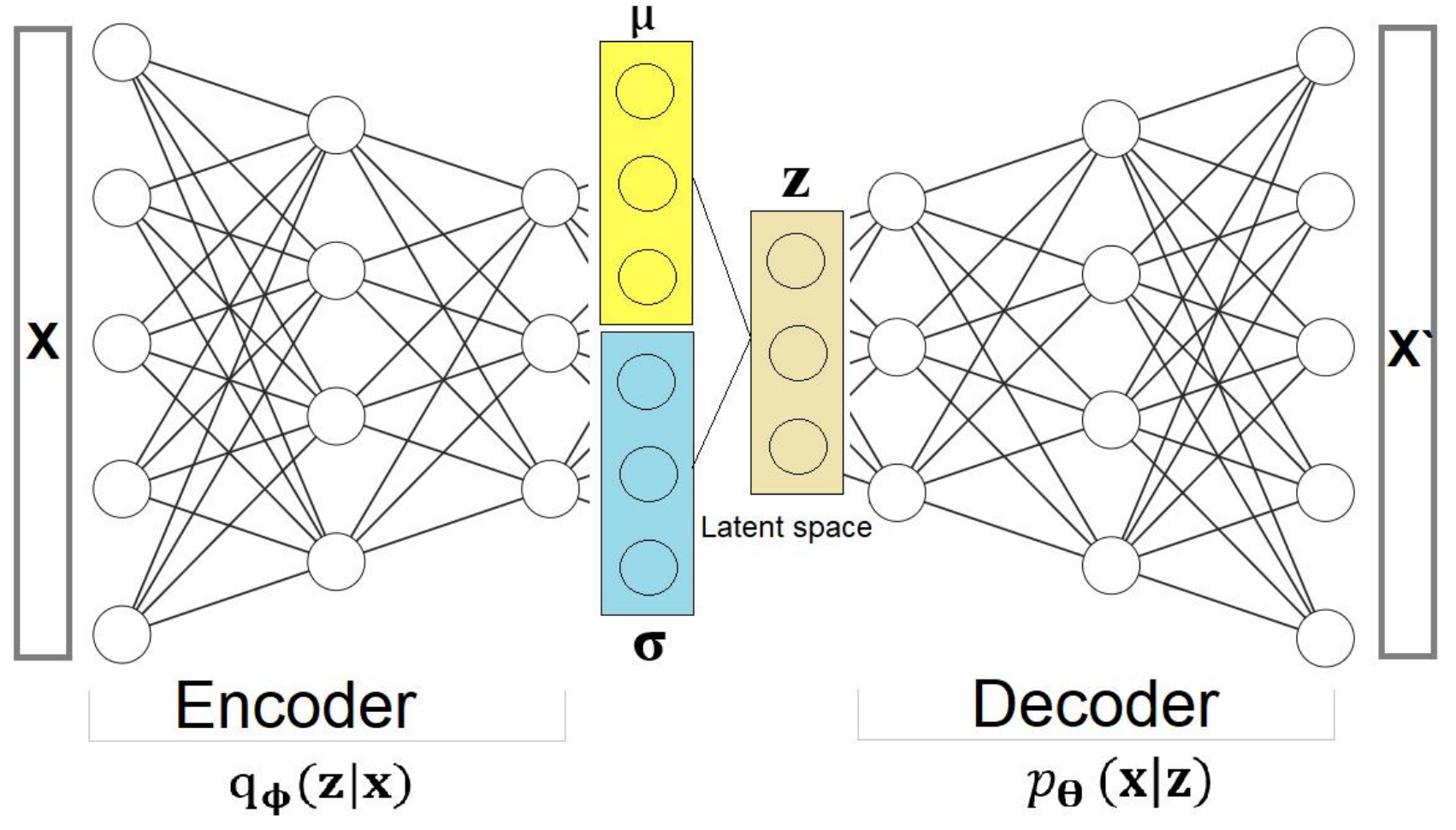}
	\caption{Illustration of the typical structure of VAEs.}
	\label{fig:VAE-structure}
\end{figure}

Unlike conventional autoencoders, a VAE model encodes input data into a distribution rather than deterministic values \cite{kingma2013auto}. Consequently, the latent space of VAEs comprises two vectors, representing the mean value and standard deviation of the encoded multi-dimensional distribution, as shown in Figure \ref{fig:VAE-structure}. This method establishes a structured latent space suitable for effective data generation. For new samples generation, the decoder takes random input vector from the encoded distributions in the latent space and generates new samples. 

Define $\mathbf{x}$ as the set of observed variables with a distribution denoted as $p(\mathbf{x})$, and $\mathbf{z}$ as the set of latent variables with a joint distribution $p(\mathbf{z},\mathbf{x})$. We can compute the conditional probability $p(\mathbf{z}|\mathbf{x})$ using the Bayes' rule:

\begin{equation}\label{eqn:Bayes-rule}
	p(\mathbf{z}|\mathbf{x}) = \frac{p(\mathbf{x}|\mathbf{z}) p(\mathbf{z})}{p(\mathbf{x})}
\end{equation}

However, the computation of $p(\mathbf{z}|\mathbf{x})$ is challenging due to the intractable integral in the marginal likelihood term (also called evidence), $p(\mathbf{x}) = \int p(\mathbf{x}|\mathbf{z}) p(\mathbf{z}) d\mathbf{z}$. Therefore, the posterior distribution $p(\mathbf{z}|\mathbf{x})$ is very difficult to learn. To address this issue, we approximate $p(\mathbf{z}|\mathbf{x})$ with a tractable distribution $q(\mathbf{z}|\mathbf{x})$ using variational inference (VI) \cite{blei2017variational}. VI reformulates the inference problem as an optimization task. It approximates the complex distribution $p(\mathbf{z}|\mathbf{x})$ with a simpler distribution, often chosen as Gaussian, and minimizes the difference between them, measured using the Kullback-Leibler (KL) divergence as follows:

\begin{align} \label{eqn:KL-divergence}
	\mathcal{D}_{\text{KL}} ( q_{\bm{\phi}}(\mathbf{z}|\mathbf{x}) \Vert p_{\bm{\theta}}(\mathbf{z}|\mathbf{x}) ) &= \sum_\mathbf{z} q_{\bm{\phi}}(\mathbf{z}|\mathbf{x}) \log \left( \frac{ q_{\bm{\phi}}(\mathbf{z}|\mathbf{x}) }{ p_{\bm{\theta}}(\mathbf{z}|\mathbf{x})} \right ) \\
	&= \mathbb{E}_{\mathbf{z} \sim q_{\bm{\phi}}(\mathbf{z}|\mathbf{x}) } \left[ \log(q_{\bm{\phi}}(\mathbf{z}|\mathbf{x}) ) - \log( p_{\bm{\theta}}(\mathbf{z}|\mathbf{x}) ) \right]
\end{align}

In this context, the approximation of the encoder network is denoted as $q_{\bm{\phi}}(\mathbf{z}|\mathbf{x})$, and the decoder network as $p_{\bm{\theta}}(\mathbf{x}|\mathbf{z})$. Table \ref{table:VAEs_symbols_definitions} provides a list of the symbols used and their definitions. The parameters $\bm{\phi}$ and $\bm{\theta}$ represent network parameters optimized during VAE model training. Substituting Equation (\ref{eqn:Bayes-rule}) into Equation (\ref{eqn:KL-divergence}) and simplifying, we derive the loss function as follows:

\begin{equation} \label{eqn:loss-function-derivation-vae}
	\begin{aligned}
		\mathcal{L} (\bm{\theta},\bm{\phi})&= - \mathbb{E}_{\mathbf{z} \sim q_{\bm{\phi}}(\mathbf{z}|\mathbf{x}) } \left[ \log(  p_{\bm{\theta}}(\mathbf{x}|\mathbf{z}) )\right] + \mathcal{D}_{\text{KL}} ( q_{\bm{\phi}}(\mathbf{z}|\mathbf{x}) \Vert p_{\bm{\theta}}(\mathbf{z}) ) \\
		&=  -\log ( p_{\bm{\theta}}(\mathbf{x}) ) + \mathcal{D}_{\text{KL}} ( q_{\bm{\phi}}(\mathbf{z}|\mathbf{x}) \Vert p_{\bm{\theta}}(\mathbf{z}|\mathbf{x}) )
	\end{aligned}
\end{equation}

The first term in equation (\ref{eqn:loss-function-derivation-vae}) represents \textit{the reconstruction loss}, which measures the dissimilarity between the input data and the reconstructed data. It encourages the VAE model to generate data that closely resembles the training data. The second term is known as \textit{the regularization term}. This term imposes structure and regularization on the latent space, ensuring that each vector within the latent space corresponds to meaningful generated samples.

\begin{table}[htbp] 
	\footnotesize
	\captionsetup{justification=centering}
	\caption{Definitions of mathematical symbols for VAEs.}
	\label{table:VAEs_symbols_definitions}
	\centering
	\begin{tabular}{cl}
		\toprule
		Symbols & Meanings \\ 
		\midrule
		$\mathbf{z}$  & Latent variables  \\
		$\mathbf{x}$  & Samples from real dataset    \\
		$\bm{\phi}, \bm{\theta}$ & Parameters for distributions or neural networks \\
		$p_{\bm{\theta}} (\mathbf{z})$ &  Distribution of $\mathbf{z}$\\
		$p_{\bm{\theta}} (\mathbf{x})$ &  Distribution of $\mathbf{x}$\\
		$p_{\bm{\theta}} (\mathbf{x}|\mathbf{z})$ & Decoder network \\ 
		$p_{\bm{\theta}} (\mathbf{z}|\mathbf{x})$ & Encoder network\\ 
		$q_{\bm{\phi}} (\mathbf{z}|\mathbf{x})$ & The VI approximation of the Encoder network\\
		$\mathcal{D}_{\text{KL}}$ & The Kullback–Leibler divergence\\
		$\mathcal{L} (\bm{\theta},\bm{\phi}) $ &The loss function to train the VAE model\\ 
		\bottomrule
	\end{tabular}
\end{table}

The loss function is commonly known as the \textit{variational free energy}, or \textit{expected lower bound}, which is the negative of the evidence lower bound. The term ``lower bound'' is used because the KL divergence is inherently non-negative. Consequently, when training VAEs, the act of minimizing this loss is equivalent to maximizing the lower limit of the probability for generating real data samples. Once the model has been trained, the decoder alone can be used to generate synthetic samples.

\subsection{UQ with Conformal Prediction}

In this work, we will use CP to evaluate the CIs of the DNN models predictions. Given a fixed confidence level, the DNN model that gives smaller CI width in its predictions is believed to be better. CP \cite{vovk2005algorithmic} \cite{shafer2008tutorial} is a statistical framework designed to quantify uncertainty in the predictions of any deterministic algorithm. Unlike traditional statistical methods, which often provide a point prediction with an associated standard deviation, CP produces prediction sets or CIs for regression models, ensuring that predictions capture the true value of the response variable with a specified interval. This paper focuses on CP for ML regression tasks, following the definitions and methods outlined in \cite{tibshirani2023conformal}. The objective of CP is to construct prediction or confidence sets that offer reliable uncertainty estimates. CP achieves this by computing residuals (the differences between observed and predicted values) in a way that treats both the training and test data symmetrically. This symmetry is essential because it guarantees that the residuals satisfy the property of exchangeability, a key condition for the validity of conformal predictions. Exchangeability means that the joint distribution of the responses remains unchanged under permutations of the data points, ensuring the robustness and correctness of the prediction sets. This property allows CP to provide prediction intervals with valid coverage guarantees, i.e., the true response value will lie within the predicted interval with a specified probability, such as 95\%.

Firstly, we divide the dataset for a given problem into independent training set $D_{1}$ with $n_{1}$ number of samples and testing set $D_{2}$ with $n_{2}$ samples, this approach is called split CP. Next, we choose a deterministic ML model or point predictor to train on  $(X_{i}, Y_{i}), \: i \in D_{1}$. The following steps illustrates split CP implementation for regression algorithms:  

\begin{enumerate}
    \item Compute the residual as absolute error on test set
    \begin{equation}\label{eqn:CP-residual}
        R_{i} = |Y_{i} - \Hat{f}_{n_{1}}(X_{i})|, \: i \in D_{2}
    \end{equation}

    \item Compute the conformal quantile of $R_{i}$:
    \begin{equation}\label{eqn:CP-conformal-quantile}
        \Hat{q}_{n_{2}} = [(1 - \alpha) (n_{2} + 1)] \: \text{smallest} \: \text{of} \: R_{i}, \: i \in D_{2}
    \end{equation}

    \item The prediction or confidence set is computed as:
    \begin{equation}\label{eqn:CP-CI}
        \Hat{C}_{n} (x) = [\Hat{f}_{n_{1}}(x) - \Hat{q}_{n_{2}},  \Hat{f}_{n_{1}}(x)+ \Hat{q}_{n_{2}}]
    \end{equation}
\end{enumerate}
where $Y_{i}$ is the true response values, $\Hat{f}_{n_{1}}(x)$ is the prediction of the ML model on $D_{1}$, and $\alpha$ is the confidence level. The direct results from split CP we get is:
\begin{equation}\label{eqn:CP-direct-result}
    \mathbb{P}\left( Y_{n+1} \in \Hat{C}_{n} (X_{n+1}) \bigg| (X_{i}, Y_{i}), \: i \in D_{1} \right) \in \left[ 1 - \alpha, \: 1 - \alpha + \frac{1}{n_{2} + 1} \right]
\end{equation}
where $\mathbb{P}$ is a probability that a new response point $Y_{n+1}$ lies within the confidence interval $\Hat{C}_{n}$ of the new feature point $X_{n+1}$   One key advantage of split CP is that it accepts any error metric as residual function or sometimes refereed as conformity scores in CP literature. Let $E (x,y) = E\left( (x,y);\Hat{f}_{n_{1}}\right)$ assigns a residual function to the point $(x,y)$ based on $\Hat{f}_{n_{1}}$. The generalized residual function is defined as:
\begin{equation}\label{eqn:CP-general-residual}
    R_{i} = E\left( X_{i}, Y_{i} \right), \: i \in D_{2}
\end{equation}
and the CI will be:
\begin{equation}\label{eqn:CP-general-CI}
            \Hat{C}_{n} (x) = \left\{ y: E(x,y) \leq [(1-\alpha)(n_{2} + 1)] \: \text{smallest of} \: R_{i}, \: i \in D_{2} \right\}
\end{equation}
If the performance metric is positively-oriented (higher values indicate better model performance), the metric is negated before passing to CI computation. The CI is written as:
\begin{equation}\label{eqn:CP-positive-metrics-CI}
            \Hat{C}_{n} (x) = \left\{ y: E(x,y) \geq [\alpha (n_{2} + 1)] \: \text{smallest of} \: R_{i}, \: i \in D_{2} \right\}
\end{equation}
Keeping the performance metric generic, split CP set can be written in other formulations rather than the one defined in equation (\ref{eqn:CP-general-CI}):
\begin{equation}\label{eqn:CP-weights-CIs}
         \Hat{C}_{n} (x) = \left\{ y: E(x,y) \leq \: \text{Quantile} \: \left(\frac{[(1-\alpha)(n_{2} + 1)]}{n_{2}} ; \frac{1}{n_{2}} \sum_{i \in D_{2}} \delta_{R_{i}}\right) \right\}\\
\end{equation}
\begin{equation}\label{eqn:CP-cdf-CIs}
             \Hat{C}_{n} (x)= \left\{y: \frac{1}{n_{2}} \sum_{i \in D_{2}} 1\{R_{i} \leq E(x,y)\} \leq \frac{[(1-\alpha)(n_{2} + 1)]}{n_{2}} \right\}
\end{equation}
The second formulation in equation (\ref{eqn:CP-weights-CIs}) follows the structure ``test score $\leq$ adjusted quantile''. This structure is particularly useful for extending CP to incorporate weights when the scores are no longer exchangeable. The third formulation in equation (\ref{eqn:CP-cdf-CIs}) is defined in terms of the empirical cumulative distribution function (CDF) of $R_{i}, \: i \in D_{2}$. It follows the structure ``CDF evaluated at test score $\leq$ adjusted level'', which is beneficial when considering auxiliary randomization schemes.

\subsection{UQ with BNNs}

\begin{figure}[!htb]
	\centering
	\captionsetup{justification=centering}
	\includegraphics[width=0.8\textwidth]{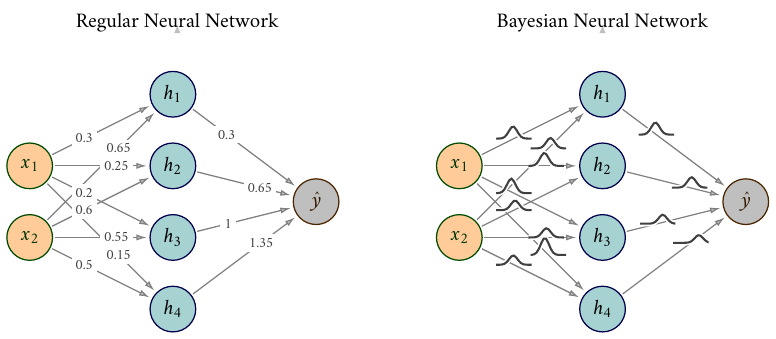}
        \vspace*{10pt}
	\caption[]{Bayesian neural network illustration.}
	\label{fig:demo_regular_vs_Bayes_NN}
\end{figure}

A regular DNN contains of two types of learnable parameters: weights and biases. During training, these parameters are optimized to specific values to achieve the highest possible accuracy. Consequently, the predictions for a given input are always deterministic. In contrast, a BNN \cite{ghahramani2016history} \cite{goan2020bayesian} is a DNN that introduces probabilistic treatment to these parameters. Instead of fixed values, the weights and biases follow probability distributions. During training, prior distributions are initially assigned to the neural network (NN) parameters, and using Bayes' theorem, these priors are updated with training data to obtain posterior distributions. At the prediction stage, the BNN is evaluated multiple times for the same input, sampling its parameters from the learned posterior distributions each time. This process generates a distribution of predictions for the same input, providing a measure of predictive uncertainty. In essence, a BNN extends a standard DNN by applying a probabilistic framework to its parameters, necessitating specialized training methods. Figure \ref{fig:demo_regular_vs_Bayes_NN} illustrates the differences between a regular neural network and a BNN.

DNNs contain a large number of parameters, making the inference of their posterior distributions a complex and computationally expensive task. To address this, several methods have been developed for Bayesian inference in NNs, including sampling-based techniques such as Markov Chain Monte Carlo (MCMC) \cite{neal2012bayesian} and optimization-based approaches like variational Bayesian inference (VI) \cite{blei2017variational}\cite{tzikas2008variational}. VI methods are particularly advantageous due to their faster convergence, making them well-suited for large-scale NNs. In this study, we utilized VI to train the BNN. The bias parameters were treated as deterministic, as NN predictions are generally more sensitive to weights than to biases. This is because the bias term is added to the product of weights in the activation function from the preceding hidden layer, resulting in weights having a more substantial influence on DNN predictions than biases.

For the BNN, a probabilistic model is assumed, where the weights are learned through Maximum Likelihood Estimation. During training, the posterior weights $(\mathbf{w})$ are determined using Bayes' rule, given a specific training dataset $(\mathcal{D})$:
\begin{equation}
	P (\mathbf{w} | \mathcal{D})  =  \frac{P (\mathcal{D} | \mathbf{w}) \cdot P (\mathbf{w})}{P (\mathcal{D})}
\end{equation}
where $P (\mathbf{w})$ is the prior distribution for $\mathbf{w}$ and it is assumed to be certain non-informative distribution, $P (\mathcal{D} | \mathbf{w})$ is the likelihood function, and $P (\mathbf{w} | \mathcal{D})$ is the posterior distribution for $\mathbf{w}$. Prior and posterior represent our knowledge of $\mathbf{w}$ before and after observing $\mathcal{D}$, respectively. $P (\mathcal{D})$ does not contain $\mathbf{w}$ so it is usually treated as a normalizing constant. It is sometimes referred to as the evidence term. When making predictions at a test data $\mathbf{x}^{*}$, the predictive distribution of the output $\mathbf{y}^{*}$ is given by:
\begin{equation}
	P (\mathbf{y}^{*} | \mathbf{x}^{*})  =  \mathbb{E}_{P (\mathbf{w} | \mathcal{D})} \left[  P (\mathbf{y}^{*} | \mathbf{x}^{*}, \mathbf{w})  \right]
\end{equation}
where the expectation operator $\mathbb{E}_{P (\mathbf{w} | \mathcal{D})}$ means we need to integrate over $P (\mathbf{w} | \mathcal{D})$. The term $ P (\mathbf{y}^{*} | \mathbf{x}^{*}, \mathbf{w})$ represents the probability of the prediction at a test point $\mathbf{x}^{*}$ and the posteriors of the weights. Each possible configuration of the weights, weighted according to the posterior distribution $P (\mathbf{w} | \mathcal{D})$, makes a prediction about $\mathbf{y}^{*}$ given $\mathbf{x}^{*}$. This is why taking an expectation under the posterior distribution on weights is equivalent to using an ensemble of an infinite number of neural networks. Unfortunately, such expectation operation is intractable for neural networks of any practical size, due to a large number of parameters as well as the difficulty to perform exact integration. This is the main motivation to use a variational approximation for $P (\mathbf{w} | \mathcal{D})$. VI methods are a family of techniques for approximating intractable integrals arising in Bayesian inference and ML. VI is used to approximate complex posterior probabilities that are difficult to evaluate directly as an alternative strategy to MCMC sampling. An alternative variational distribution is proposed to approximate $P (\mathbf{w} | \mathcal{D})$, it consists of a distribution set whose  parameters are optimized using the KL divergence method. More details on BNN training and demonstration examples for nuclear engineering problems can be found in our previous work \cite{xie2024functional, yaseen2023quantification, moloko2023prediction}. Interested readers can refer to \cite{blei2017variational} \cite{blundell2015weight} for more detailed theory of VI and BNN.

\section{Results}
\label{section:Results}

\begin{figure}[!htb]
	\centering
	\captionsetup{justification=centering}
	\includegraphics[width=0.8\textwidth]{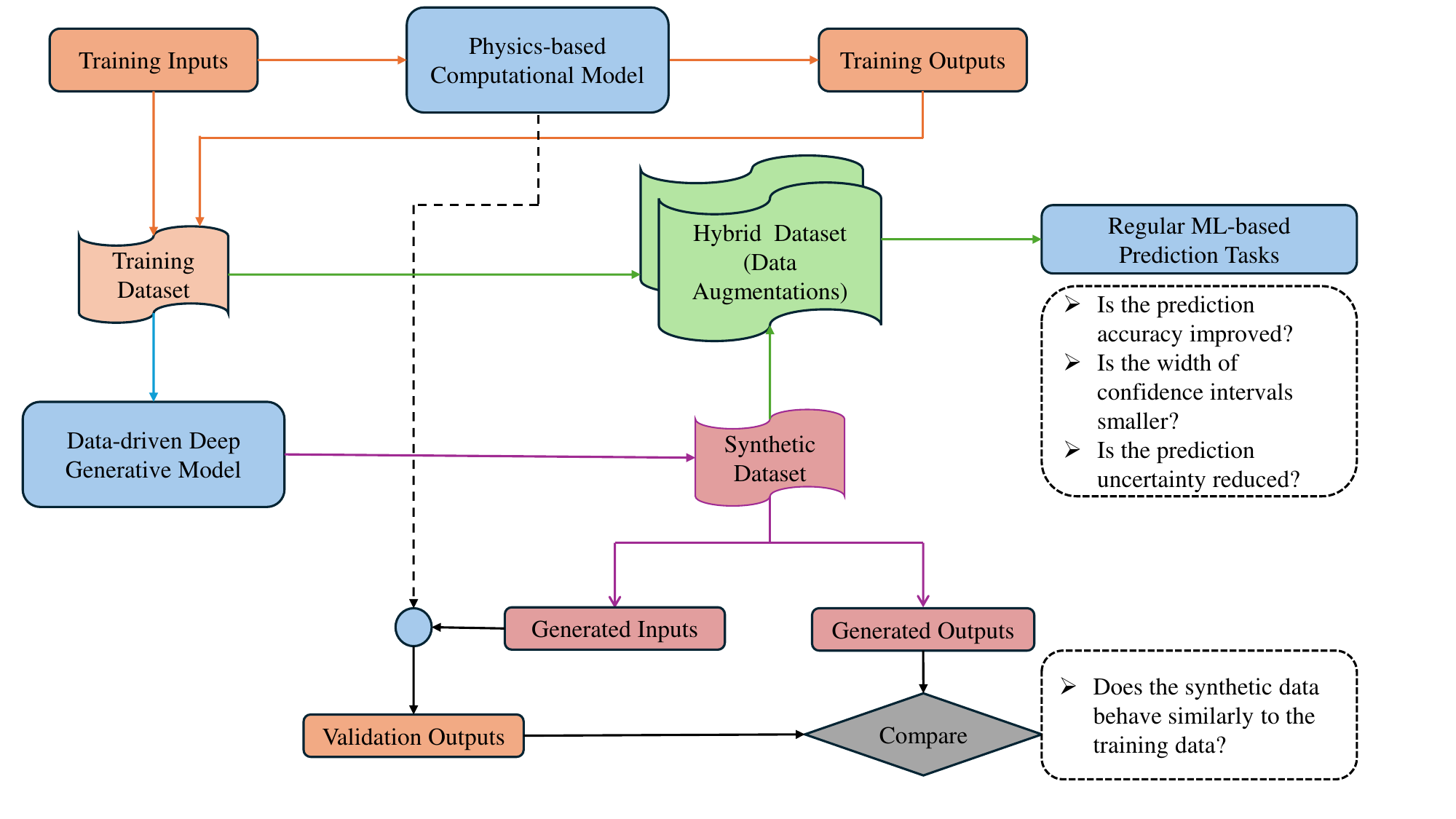}
        \vspace*{10pt}
	\caption[]{The workflow to evaluate the performance of data augmentation using the synthetic samples generated by VAEs.}
	\label{fig:workflow-diagram}
\end{figure}

Data augmentation was performed to expand the size of the original training dataset. Figure \ref{fig:workflow-diagram} illustrates the workflow of data augmentation. In our previous work \cite{alsafadi2023deepNED}, a VAE model was trained using the original training dataset and used to generate 500 synthetic samples. These samples were validated by comparing them with TRACE output for void fraction values. These synthetic samples were found to agree well with TRACE results, which answered the question ``Does the synthetic data behave similarly to the training data'' in Figure \ref{fig:workflow-diagram}. In this work, these 500 synthetic samples will first be used to assess the impact of data augmentation on the prediction accuracy of DNN models. Secondly, CP will be employed to compute prediction CIs for DNNs models. The width of these intervals will serve as a performance metric to evaluate the impact of incorporating synthetic data for training. Finally, the effect of data augmentation on uncertainty estimation through BNN will be analyzed. These three steps will quantitatively answer the two questions in the top right of Figure \ref{fig:workflow-diagram}.

\subsection{Effect of data augmentation on performance of DNNs}
\label{subsec:ANN-results}

To assess the impact of data augmentation on the performance of the DNN model, we compared a sequence of DNN models, trained using different amount of data. We began by training a baseline DNN model using the original training dataset without any synthetic samples. Subsequently, we created five additional training datasets by incrementally expanding the original training dataset by 100 random synthetic samples each time, resulting in ``hybrid datasets'' (TRACE simulation data plus VAE-generated synthetic samples) that contain 300, 400, 500, 600, and 700 samples, respectively. These datasets were then used to train five separate DNN models, each of which underwent optimization through hyperparameter tuning. 

\begin{figure}[htb]
	\centering
	\begin{subfigure}{0.495\textwidth} 
		\centering
		\includegraphics[width=0.9\linewidth]{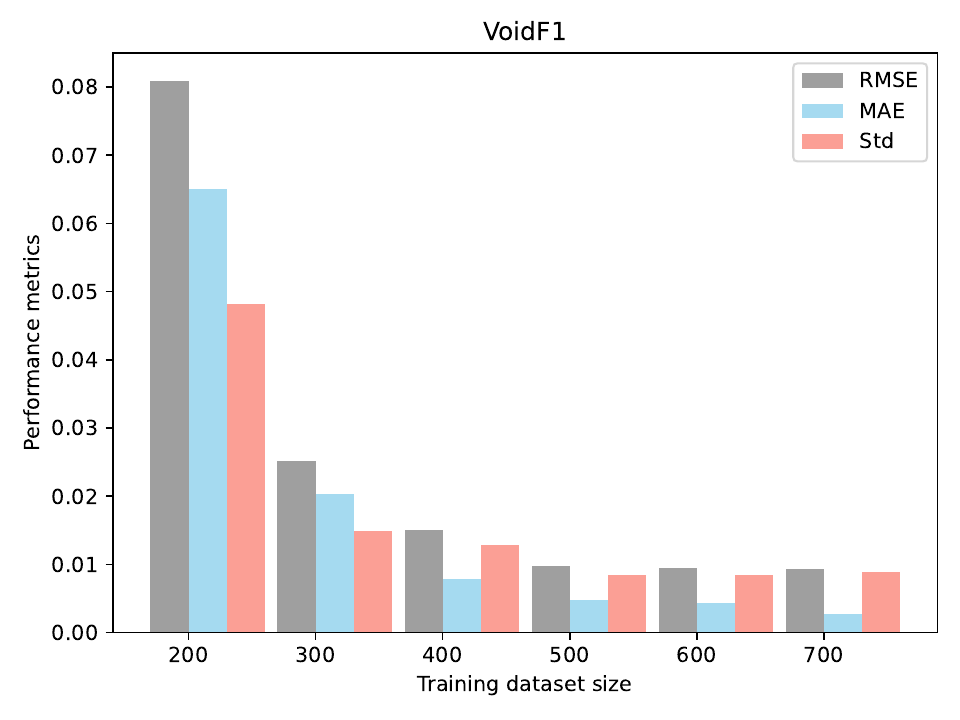}
	\end{subfigure}%
	\hspace{-2em} 
	\begin{subfigure}{0.495\textwidth}
		\centering
		\includegraphics[width=0.9\linewidth]{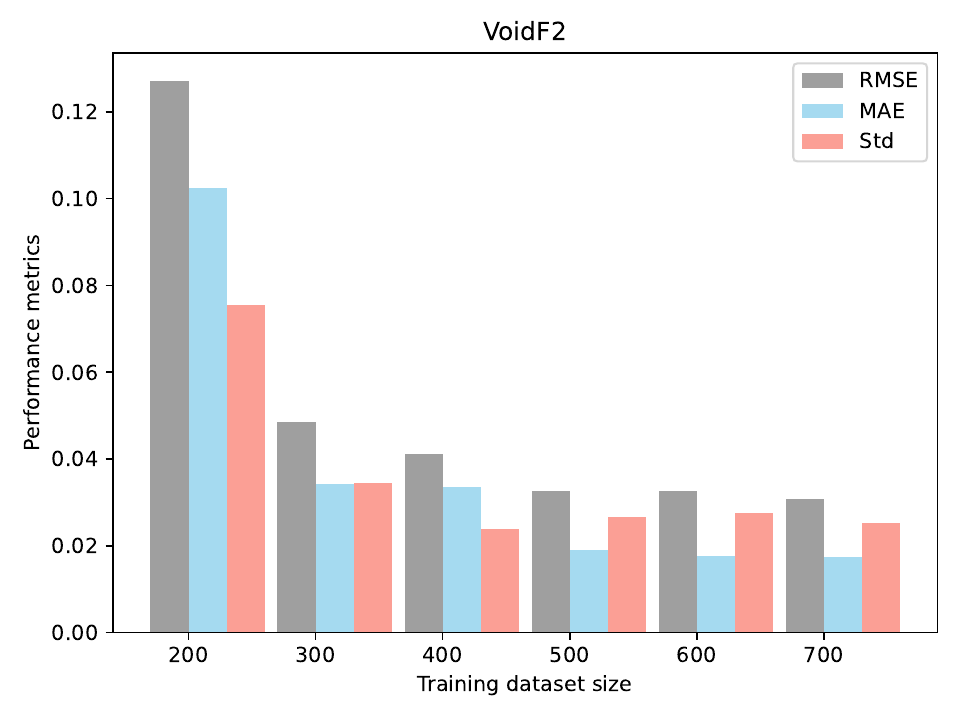}
	\end{subfigure}
	\vspace{0.5em} 
	\begin{subfigure}{0.495\textwidth}
		\centering
		\includegraphics[width=0.9\linewidth]{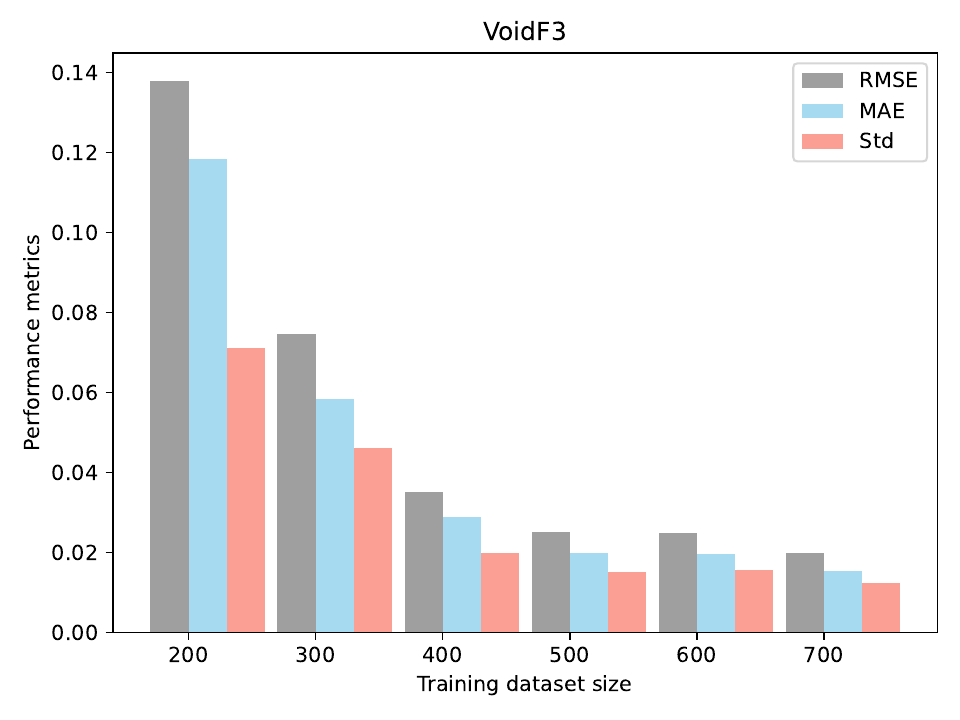}
	\end{subfigure}%
	\hspace{-2em} 
	\begin{subfigure}{0.495\textwidth}
		\centering
		\includegraphics[width=0.9\linewidth]{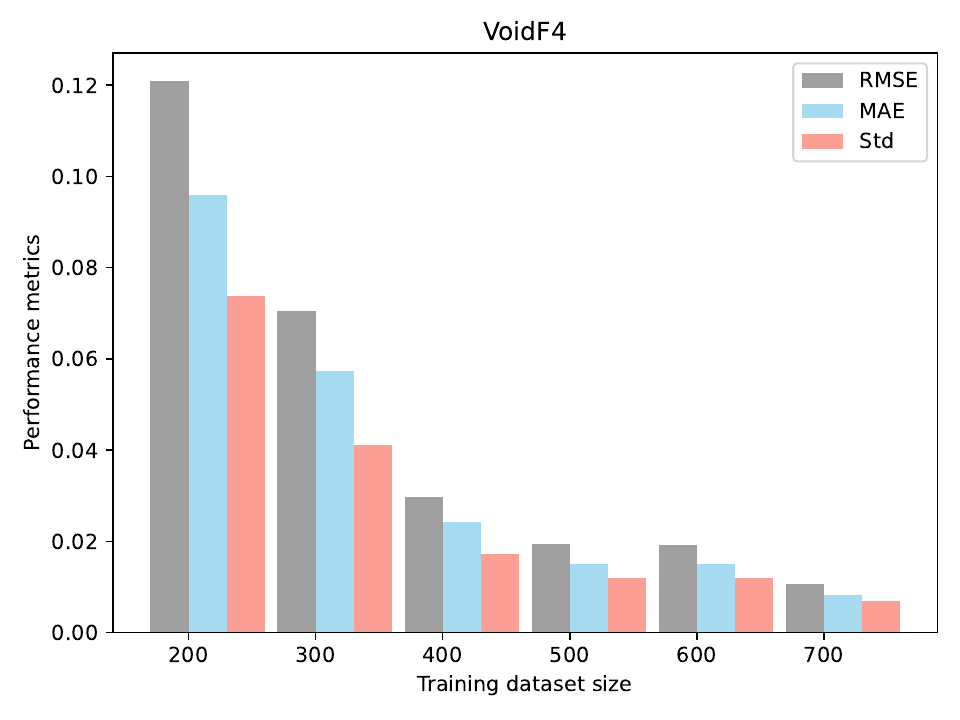}
	\end{subfigure}
	\caption{Performance comparison of DNNs in terms of their RMSE, MAE and the standard deviation of absolute errors on the test dataset, using different training dataset sizes.}
	\label{fig:Performance-of-DNN}
\end{figure}

To compare the predictive accuracy of the DNN models trained with the six different datasets, we computed the absolute prediction errors using the standalone test dataset. In each instance, the testing dataset consisted of the same 200 samples from TRACE simulations, which were unseen during the training of both the VAEs generative model and the DNN predictive models. We calculated the mean absolute error (MAE), the root mean squared error (RMSE) and the standard deviations of the errors to assess the model's performance. The MAEs, RMSEs and the standard deviations for the four void fractions ($\texttt{VoidF1}, \texttt{VoidF2}, \texttt{VoidF3}, \texttt{VoidF4}$) are shown in Figure \ref{fig:Performance-of-DNN}. 

When comparing the MAEs and RMSEs across the six models, it is shown in Figure \ref{fig:Performance-of-DNN} that these values generally decrease as the training dataset size increases by adding more synthetic data samples. This trend holds true for all four void fraction outputs. It is also the case when comparing the standard deviation of errors, where it is noticeable that they are significantly reduced as the dataset size increases, reaching its lowest when employing the dataset with the largest number of samples. 

In some cases, the reduction in error is less pronounced when adding 100 points to the dataset. For instance, when comparing the 500 and 600 training datasets for $\texttt{VoidF1}$, the reduction in error is relatively small. However, a substantial difference can still be observed when comparing the model trained with the original dataset and the one with 700 samples. In contrast, at other void fraction outputs, the error values initially started higher and reduced significantly to smaller values. 
Nevertheless, all four outputs ultimately reached similarly small error values. These findings indicate that augmenting the training dataset has improved the DNN model's predictive performance.

\subsection{Effect of data augmentation on DNNs prediction CIs}
\label{subsec:DNN-CI-CP-results}

\begin{figure}[!htb]
	\centering
	\captionsetup{justification=centering}
	\includegraphics[width=0.75\textwidth]{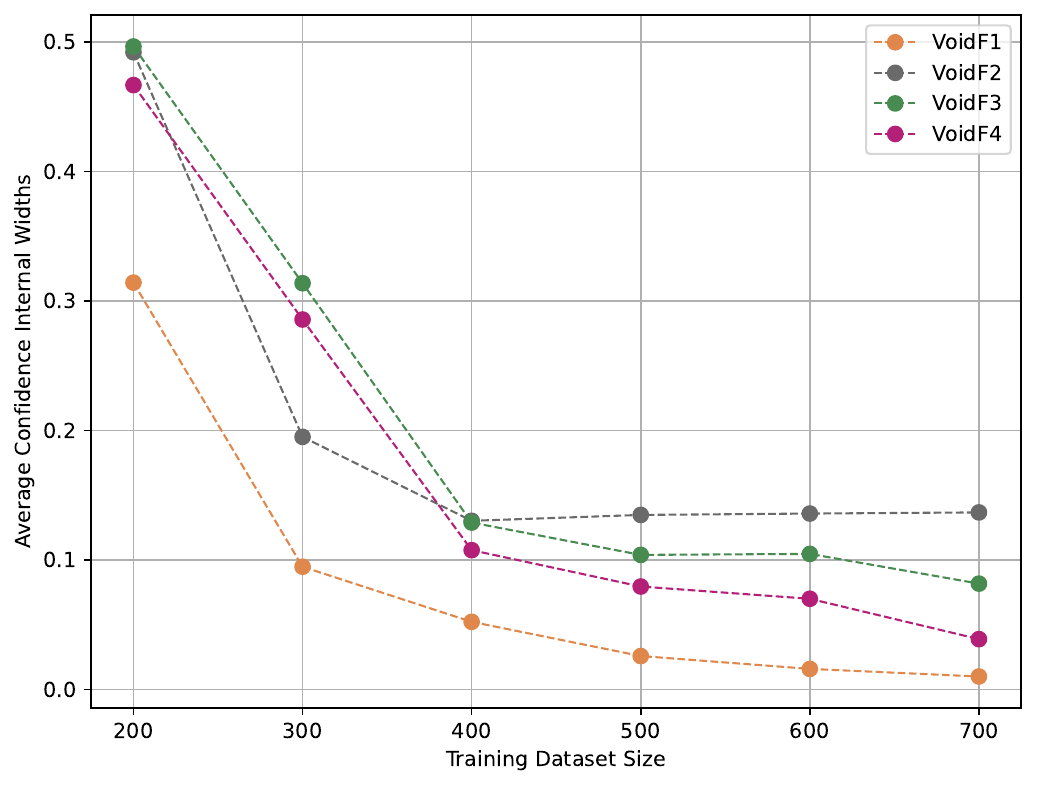}
	\caption[]{Effect of data augmentation on DNNs CIs average width.}
	\label{fig:UQ-DNN-CI-CP}
\end{figure}

Using the CP method, we computed CIs for the deterministic predictions of DNNs without the need to train a probabilistic ML model. In this study, we have chosen a confidence level of 95\%, the corresponding CI means that the DNN model has a 95\% prediction confidence that the true output value at the test sample lies within this CI. To evaluate the impact of adding synthetic data, we calculated the average CI width, which represents the mean difference between the upper and lower CI limits across the test dataset. Figure \ref{fig:UQ-DNN-CI-CP} illustrates how this metric changes with the introduction of new data. Notably, the CI width decreases for all void fraction outputs as synthetic data is added. This reduction corresponds to the lower test error observed in the DNN predictions, as depicted in Figure \ref{fig:Performance-of-DNN}, indicating that the range within which the true void fraction value lies has become narrower.

For all the four void fraction outputs, the CI widths significantly decrease until the dataset size reaches 400, after which further changes are minimal. This plateau can be attributed to the slight variation in RMSE and MAE, suggesting that the DNN has achieved a balance between bias and variance, with limited performance gains from additional data. The improved DNN performance, coupled with smaller CIs, suggests that augmenting data with deep generative models enhances the model’s deterministic regression capabilities and confidence.

\subsection{Effect of data augmentation on UQ}

\begin{figure}[!htb]
	\centering
	\captionsetup{justification=centering}
	\includegraphics[width=0.95\textwidth]{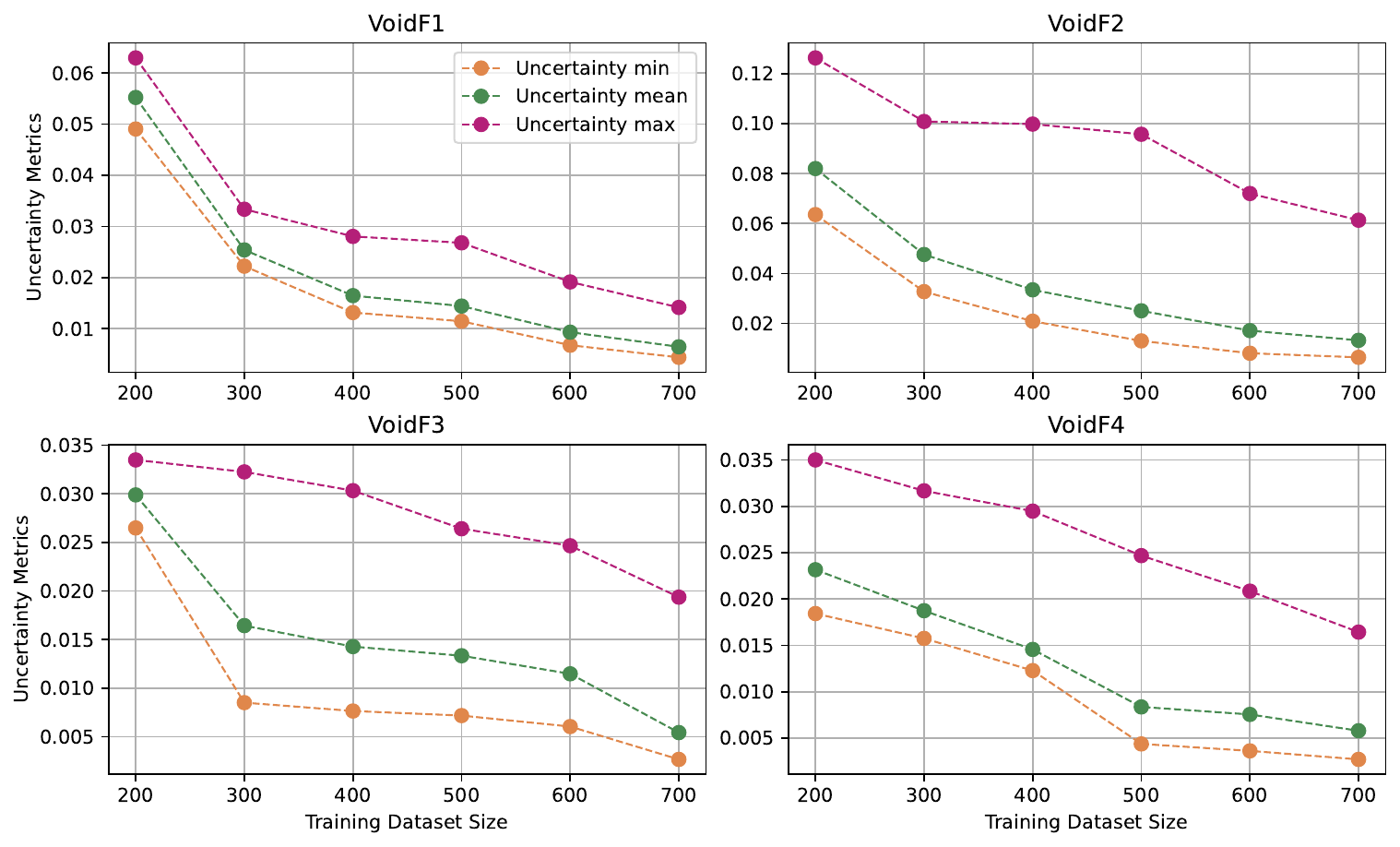}
	\caption[]{Effect of data augmentation on uncertainty estimated using BNN.}
	\label{fig:UQ-DNN-BNN}
\end{figure}

To evaluate the predictive uncertainty of the DNN models, we trained BNNs using the VI method. For each BNN model, we incrementally expanded the synthetic dataset by 100 data points. For a specific input, the BNN model produces both a mean value, representing the averaged prediction of the output, and a standard deviation, indicating the prediction's uncertainty. In Figure \ref{fig:UQ-DNN-BNN}, we illustrate the minimum, mean, and maximum prediction uncertainties for test samples versus various dataset sizes used to train the BNN models. Notably, as we increase the number of samples of the ``hybrid dataset'' (simulation samples plus synthetic samples), the prediction uncertainty decreases consistently for the optimized BNN model. This reduction in uncertainty suggests that the neural network is making more confident predictions with smaller uncertainty.

\begin{figure}[!htb]
	\centering
	\captionsetup{justification=centering}
	\includegraphics[width=0.75\textwidth]{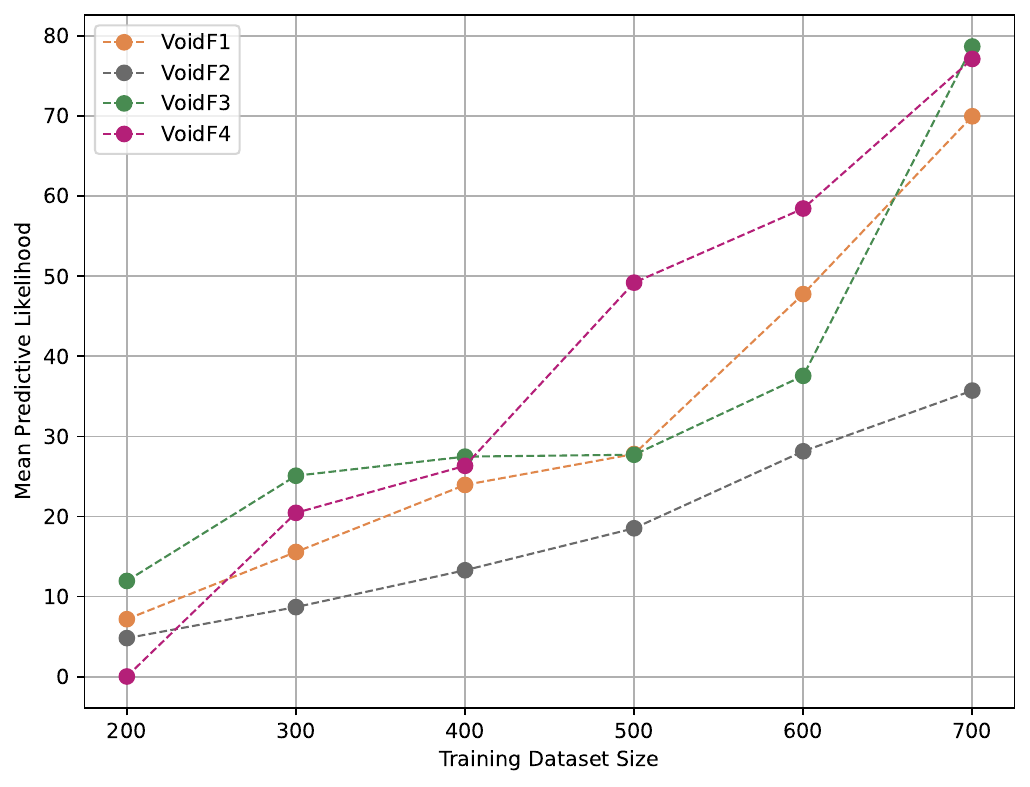}
	\caption[]{The MPL metric vs. training dataset size.}
	\label{fig:UQ-BNN-MPL}
\end{figure}

To further evaluate the predictive capacity we also used a metric called \textit{mean predictive likelihood (MPL)}, which measures the average likelihood of observed data points under the model's predictive distribution. High MPL values can be interpreted as the test data being highly probable under the predictive distribution. Thus indicating that the ML model is making predictions that are consistent with the data and its associated uncertainty. MPL is given by:
\begin{equation}
	\text{MPL} = \frac{1}{N_{\text{test}}} \sum^{N_{\text{test}}}_{i = 1} \mathcal{N} (\mu (x_i), \sigma^2 (x_i)) [u_i]
\end{equation}
where $N_{\text{test}}$ is the number of test samples ($N_{\text{test}}=200$ in this work), $\mathcal{N} (\mu (x_i), \sigma^2 (x_i)) [u_i]$ denotes the probability density function value of the tested data point $u_i$ under a Gaussian distribution with mean value $\mu (x_i)$ and variance $\sigma^2 (x_i)$. Gaussian distribution is the most widely used and natural choice for this purpose. This assumption has also been validated by plotting the testing samples outputs together, which closely follow a Gaussian distribution. The MPL values are shown in Figure \ref{fig:UQ-BNN-MPL}. It shows that larger training dataset size results in larger MPL values, indicating that the ML models are producing outputs that are more to the real outputs at the testing samples.

\section{Discussions and Conclusions}
\label{section:Conclusions}

ML models have demonstrated remarkable success in various fields, particularly in natural language processing and computer vision. These domains benefit from extensive training datasets, which contribute to the effectiveness of ML models, especially the ML models that heavily rely on ``big data''. However, in certain fields where data are scarce, the performance of ML models may reduce significantly. This is especially true in nuclear engineering, where data are often derived from high-cost and time-consuming experiments. In a previous work\cite{alsafadi2023deepNED}, we have investigated several different types of DGMs for synthetic data generation to alleviate the data scarcity issue. 

In this work, we studied the impact of data augmentation on DNN prediction accuracy, confidence interval width of DNNs computed using CP, as well as on the DNN prediction uncertainties. To achieve this, VAE models was used for generating 500 synthetic samples to expand our training dataset. Subsequently, we compared the performance of the DNN trained with the original dataset and a sequence of additional DNN models by increasing the dataset size 100 samples each time. We also assessed the uncertainties in the DNN predictions based on the different training datasets using BNNs. 

Our investigations demonstrated the effectiveness of data augmentation in improving the performance of DNN and reducing the uncertainties in their predictions, which is the primary objective of this study. Using a standalone test dataset, consistent improvements were shown in terms of the MAE, RMSE and standard deviation, as all these values decreased with increasing the training dataset size. Notably, across all void fraction outputs, the error metrics were significantly higher for the baseline DNN model, with some variations in their values and VoidF3 having the highest values. However, when the model was trained with largest dataset (700 points), the error metrics across all void fraction outputs became similarly small, with minimal deviations and VoidF1 having the smallest error metrics.

The same behaviour was observed in the CP results, where the CIs became narrower when predictions were made from the models trained on larger datasets. Comparing with the error metrics from the DNN models, we see consistency across the results. As shown in Figures \ref{fig:UQ-DNN-CI-CP}, and \ref{fig:Performance-of-DNN}, for VoidF2, models trained with dataset containing more than 400 points showed minimal changes in the CIs, which is consistent with the small changes in error metrics (MAE, RMSE, and standard deviation), for this void fraction output. This consistency holds for all other void fraction outputs as well.

Finally, the BNN results showed similar trend, where the uncertainties were reduced with increasing training dataset size. These findings illustrates the effectiveness of generative model-based data augmentation in improving the performance of ML models, resulting in more accurate predictions and reduced uncertainties. In future work, we intend to explore the effectiveness of other DGM techniques such as diffusion models for scientific data augmentation. We aim to apply these techniques to more complex experimental datasets.

\section*{Acknowledgement}

This work was funded by the U.S. Department of Energy (DOE) Office of Nuclear Energy Distinguished Early Career Program (DECP) under award number DE-NE0009467. Any opinions, findings, and conclusions or recommendations expressed in this paper are those of the authors and do not necessarily reflect the views of the U.S. DOE.

\newpage
\bibliography{./bibliography.bib}

\end{document}